%% file: main.tex
\documentclass[nonacm]{acmart}

\usepackage{graphicx}

\usepackage{xcolor}

\usepackage{amsfonts}

\usepackage{subcaption}

\usepackage{cleveref}

\title{AI Does Not Alter Perceptions of Text Messages}

\author{N'yoma Diamond}
\orcid{0000-0002-6468-1779}
\affiliation{
    \institution{University of Cambridge}
    \department{Department of Computer Science and Technology}
    \streetaddress{William Gates Building, 15 JJ Thomson Avenue}
    \city{Cambridge}
    \country{United Kingdom}
    \postcode{CB3 0FD}
}
\email{bad35@cam.ac.uk}

\authorsaddresses{}

\begin{document}

\begin{abstract}
    For many people, anxiety, depression, and other social and mental factors can make composing text messages an active challenge. To remedy this problem, large language models (LLMs) may yet prove to be the perfect tool to assist users that would otherwise find texting difficult or stressful. However, despite rapid uptake in LLM usage, considerations for their assistive usage in text message composition have not been explored. A primary concern regarding LLM usage is that poor public sentiment regarding AI introduces the possibility that its usage may harm perceptions of AI-assisted text messages, making usage counter-productive. To (in)validate this possibility, we explore how the belief that a text message did or did not receive AI assistance in composition alters its perceived tone, clarity, and ability to convey intent. In this study, we survey the perceptions of 26 participants on 18 randomly labeled pre-composed text messages. In analyzing the participants' ratings of message tone, clarity, and ability to convey intent, we find that there is no statistically significant evidence that the belief that AI is utilized alters recipient perceptions. This provides hopeful evidence that LLM-based text message composition assistance can be implemented without the risk of counter-productive outcomes.
\end{abstract}

\maketitle

\input{introduction}

\input{lit_review}

\input{methodology}

\input{results}

\input{discussion}

\input{conclusion}

\newpage
\bibliography{references}

\newpage
\appendix
\input{appendix}

\end{document}

%% file: introduction.tex
\section{Introduction}

Systems like autocorrect, autocomplete, and smart replies have become cornerstones of modern text communication. While these systems provide significant assistance day-to-day, they primarily focus on simple tasks like response prediction, spelling corrections, or sentence completion. With the sudden rise in advanced generative AI---namely large language models (LLMs) like GPT-4 and LLaMa 2---the door has opened for smarter and more capable AI assistance systems for digital writing composition. Within only one year since release, ChatGPT (GPT-3.5) has seen rapid usage growth as people have identified its practicality as a writing assistant~\cite{basicChatGPT3WritingAssistance2023,chenChatGPTOtherArtificial2023}, an education tool~\cite{alafnanChatGPTEducationalTool2023}, and a mental health assistant~\cite{carlbringNewEraInternet2023,farhatChatGPTComplementaryMental2023,garciaCanChatGPTSubstitute2023,alessaDesigningChatGPTConversational2023}, among other applications.

For many people, finding the right words to use when communicating can be an active challenge and point of stress. Anxiety, depression, and other factors can lead many to doubt their communication abilities or potentially avoid social interactions altogether. Despite the rapid uptake LLM usage, considerations for their assistance in social text communication composition (\`a la autocorrect and smart replies) have not been explored substantially. LLMs, should they be utilized, may prove to be invaluable tools towards assisting people that struggle with digital message composition. For example, AI systems may be employed to analyze and verify the sentiment or clarity of personal messages on-the-fly prior to sending. However, the implementation of such tools may itself shift how recipients perceive and interpret these messages: Sentiment towards AI-generated or AI-assisted text content---especially in the age of LLMs---is poor~\cite{hohensteinArtificialIntelligenceCommunication2023,kimAIFriendAssistant2021,liuWillAIConsole2022,jakeschAIMediatedCommunicationHow2019}. As a result, even though message composers may personally find AI-assistance beneficial~\cite{valenciaLessTypeBetter2023,fuTextSelfUsers2023}, it is possible that the actual usage of AI-assistance may instead be self-defeating as a result of negative perceptions by the recipient towards AI.

In this study we analyze how the \textit{belief} that AI assistance may have been used in message composition may alter a recipient’s perception of text messages, independent of the message’s content. That is, does believing that AI was used cause a negative (or positive) shift in perceptions of a text message’s tone/sentiment? Does this belief cause the user to think a message’s contents and information are clearer (or more unclear)? Does this belief cause the recipient to think the text composer was better or worse at conveying their desired intent and purpose? In understanding the answers to these questions we will be better able to guide the design and implementation of future assistive text composition tools for those that would benefit most from them.

%% file: lit_review.tex
\section{Literature Review}

\subsection{Impacts and Perceptions of AI-Mediated Communication}

\citet{hohensteinArtificialIntelligenceCommunication2023, mieczkowskiAIMediatedCommunicationLanguage2021} explore the impacts and perceptions of AI-generated ``smart replies'' in text communication with respect to language and social relationships. In \cite{hohensteinArtificialIntelligenceCommunication2023}, users often utilized smart replies to accelerate the process. In spite of this, when users that \textit{believe} their partner utilized smart replies perceive interactions more negatively (less cooperation, more domination, etc.). However, \textit{actual} utilization of smart replies improve these social outcomes and perceptions long-term. \cite{mieczkowskiAIMediatedCommunicationLanguage2021} specifically focuses on the effects of AI-induced positivity bias and how it impacts social interactions. Their study found that generated smart replies biased significantly towards positive sentiment (greater than that of humans). However, this bias did not translate to messages composed by users exposed to smart replies, thus potentially impacting perceptions of communications resultant to the perceived positivity of a message or series of messages~\cite{mieczkowskiAIMediatedCommunicationLanguage2021}. 

Other studies by \citet{valenciaLessTypeBetter2023,fuTextSelfUsers2023} explore how large language models (LLMs) can be used to improve AI-mediated communication (AI-MC)~\cite{hancockAIMediatedCommunicationDefinition2020}, identifying significant benefits for users of LLM-based assistance systems. In particular, \cite{fuTextSelfUsers2023} analyzes participants’ usage and perceptions of AI-MC tools utilizing LLMs for interpersonal communication and personal usage. Their analysis found that AI-MC tools significantly improved personal confidence in communication, composition capabilities, and the abilities of users to interpret communications. However, users also came across a number of issues regarding technical aspects of the AI-MC tools and unusual propensities of the LLMs to produce certain undesirable behavior, and were less desirable in casual communication contexts. Overall, participants observed reduced anxiety, improved communication quality, and other significant benefits for users of AI-MC tools~\cite{fuTextSelfUsers2023}.

\subsection{Trustworthiness of AI-Assisted Text}

\citet{liuWillAIConsole2022, jakeschAIMediatedCommunicationHow2019} analyze how readers of AI-MC communications perceive the conveyed information, specifically targeting trustworthiness. \cite{liuWillAIConsole2022} focuses on AI-assisted email composition under three perceived conditions: No AI presence, partial AI assistance, and total AI composition. This investigation identified that the more AI was stated to be present, the less trustworthy the participants perceived the email. Taking a different direction, \cite{jakeschAIMediatedCommunicationHow2019} analyzes personal profile text (e.g., as might be seen on social media like Instagram or Facebook, or business platforms like Airbnb, eBay, or Etsy) generated by AI. Their analysis identified that, when AI usage was ubiquitous (i.e., all profiles were labeled as AI-generated), profiles were trusted similarly to under the case of ubiquitous non-usage (i.e., all human-generated profiles). However, when participants were provided a mix of profiles labeled as human- or AI-generated, trust in AI-generated profiles significantly decreased~\cite{jakeschAIMediatedCommunicationHow2019}. Note that this study specifically focused on the context of Airbnb hosts, rather than social media. Because Airbnb is fundamentally a business platform involving the sale of short-term places to stay, people may perceive AI distinctly differently from how they would on social media platforms.

\subsection{Broad Perceptions of AI}

Outside of the AI-MC context, \citet{kimAIFriendAssistant2021} explore the different styles of AI assistance and how people perceive them. Specifically, they focus on the distinction between \textit{functional} and \textit{social} AI. Functional AI refers to AI systems that provide direct assistance following a request from the user, such as navigation, social media recommenders, etc. On the other hand, Social AI refers to systems that behave socially or serve social purposes, such as chatbots, or other types of social bots. In their analysis, they found that people in general perceive functional AI more positively than social AI, likely because the former is more practically useful. Note that this study was conducted prior to the release of ChatGPT (GPT-3.5) and the ensuing jump in interest in LLMs, usage of social AI, and new applications for these types of AI. As a result, discussions regarding LLMs and more state-of-the-art social AI like ChatGPT are missing, meaning the conclusions of this study may need reassessment under the new contemporary context.

%% file: methodology.tex
\section{Methodology}
\label{sec:methodology}

In this study, data is collected from volunteer participants through a Qualtrics survey. In the survey, participants are given a series of pre-composed messages to read and evaluate. The pre-composed messages are separated into 6 topic groups with 3 messages each: Advice (requests for general advice), Entertainment (discussing TV, movies, etc.), Gossip (rumors, friend-chatter, etc.), Informational (news, hobbies, etc.), Recommendation (requests for specific recommendations), and Social (hangouts/meetup requests, etc.). This produces 18 pre-composed messages, all of which are provided to participants in a fully randomized order. Alongside each message, the participant will receive either (A) a disclaimer that AI provided assistance in composition, denoted in our data by label \texttt{+}, (B) a disclaimer that AI did \textit{not} provide assistance in composition, denoted by label \texttt{-}, or (C) no disclaimer at all, denoted by label \texttt{=}. Each message within a topic group will be randomly given a different label such that every group has one message displayed with each label. That is, each message within a fixed topic group is assigned a label uniformly randomly without replacement. Thus every message has a 33\% chance of being labeled as AI-assisted, 33\% chance of being labeled as not having AI assistance, and 33\% chance of having no label at all. As a result, every participant will see 6 messages labeled as having AI assistance, 6 labeled as not having AI assistance, and 6 with no label at all, and will see each label exactly once per topic group.

For each message, the participant was asked to rate three message characteristics on Likert scales from 0 to 10: Tone, clarity, and the author's effectiveness at conveying their intent. The questions asked to the participants are as follows:
\begin{itemize}
    \item ``On a scale of 0 (very negative) to 10 (very positive), how do you perceive the tone of the message?''
    \item ``On a scale of 0 (very unclear) to 10 (very clear), how do you rate the clarity of the message?''
    \item ``On a scale of 0 (very poorly) to 10 (very well), how well do you believe the message writer conveyed their intent?''
\end{itemize}

\begin{figure}[htbp]
\centering
\begin{subfigure}{.33\linewidth}
    \centering
    \includegraphics[width=0.9\linewidth]{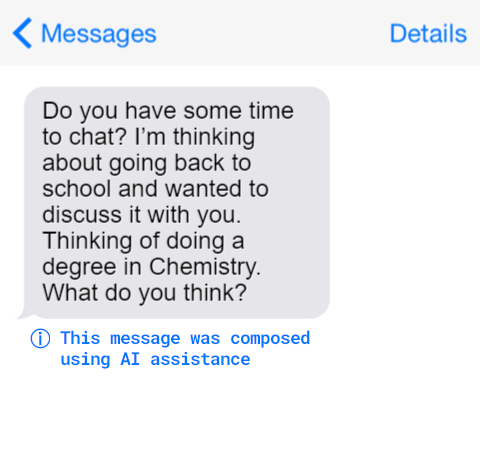}
    \caption{Advice, AI present (\texttt{+})}
\end{subfigure}%
\begin{subfigure}{.33\linewidth}
    \centering
    \includegraphics[width=0.9\linewidth]{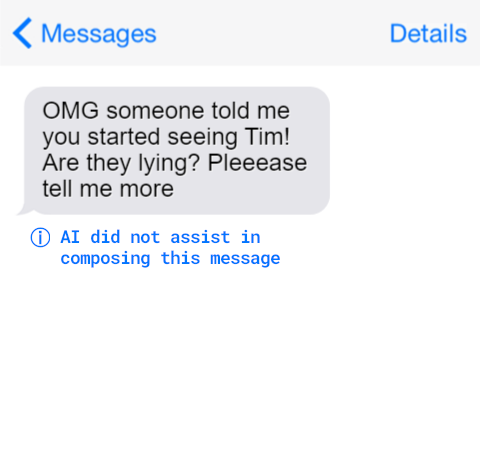}
    \caption{Gossip, AI not present (\texttt{-})}
\end{subfigure}%
\begin{subfigure}{.33\linewidth}
    \centering
    \includegraphics[width=0.9\linewidth]{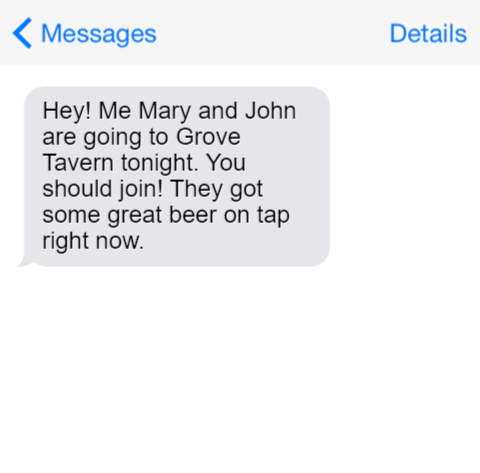}
    \caption{Social, unlabeled (\texttt{=})}
\end{subfigure}
\caption{Example messages of varying topics provided to participants}
\end{figure}

26 Participants were acquired as volunteers via word-of-mouth and social media (Discord). The people requested to participate and social media communities in which advertising occurred primarily consist of university students across a mix of Bachelor's and Master's programs, mostly consisting of STEM degrees.

To analyze our results, Tukey's Honestly Significant Difference test is utilized to evaluate the pairwise difference between the result for each question depending on label. To validate Tukey's test's assumption of homoscedasticity, Levene's test is performed and violin plots are generated to provide additional visual qualitative analysis.


%% file: results.tex
\section{Results}

Results are analyzed using Tukey’s Honestly Significant Difference test to compare if there is any statistically significant difference between the observed participant answers depending on the provided label. Tukey's test is used to correct and control for family-wise error due to performing many pairwise comparisons. To justify the usage of Tukey's test, we validate the homoscedasticity statistically using Levene's test to ensure the variances are not significantly different between the analyzed distributions. Violin plots are also used to create a visual intuition of homoscedasticity. Tests were performed in two groups: (A) Evaluation across all message topics, and (B) evaluation on within each topic grouping.

\begin{figure}[htbp]
\centering
\begin{subfigure}{.5\linewidth}
    \centering
    \includegraphics[width=\linewidth]{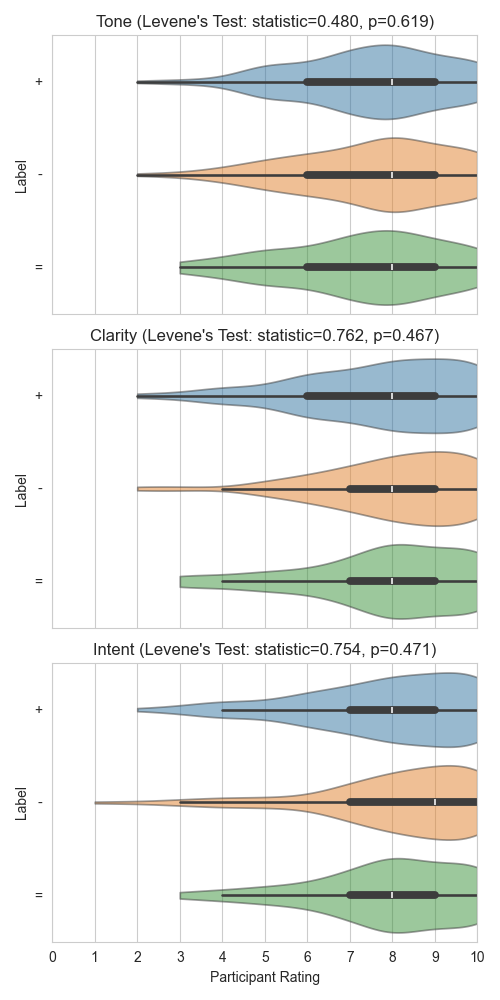}
    \caption{Violin plots with Levene's test metrics}
    \label{fig:overall_levene}
\end{subfigure}%
\begin{subfigure}{.5\linewidth}
    \centering
    \includegraphics[width=\linewidth]{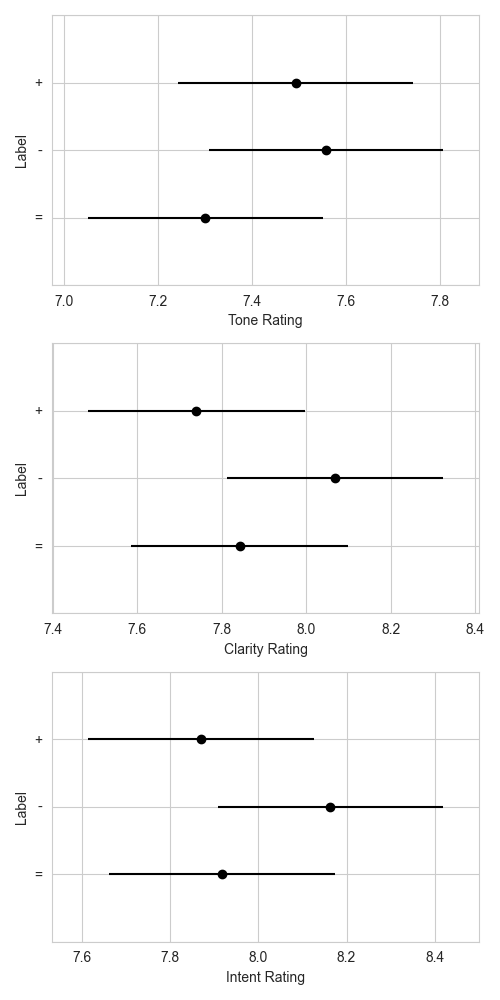}
    \caption{Interval plots for Tukey's test}
    \label{fig:overall_tukey}
\end{subfigure}
\caption{Plots for performed statistical tests}
\end{figure}

\textbf{Overall evaluation:} The performed Levene's tests (\Cref{fig:overall_levene}) show that observed results, at least when considered across all topics, display homoscedasticity. Across all questions, we fail to find statistically significant evidence that the provided label (or lack thereof) induced changes in the perceived clarity, tone, and ability to convey intent of the analyzed messages. The numeric results of the performed Tukey's tests are shown in \Cref{tab:overall_tukey_tone,tab:overall_tukey_intent,tab:overall_tukey_clarity}. It is worth noting that, even though we fail to find statistically significant evidence that the different labels induce changes in message perceptions, we see empirically that the mean perceived tone, clarity, and ability to convey intent are all lower for messages labeled as having AI assistance (\texttt{+}) compared to messages labeled as \textit{not} having AI assistance (\texttt{-}). Similarly, messages with no label (\texttt{=}) also have lower mean perceptions compared to those labeled as not having AI assistance (\texttt{-}). Thus an effect still may be present, despite our failure to find significant to validate it. This suggests that a larger sample size and/or better granularity in participant quantifications of tone, clarity, and ability to convey intent may draw out this effect.

\begin{table}[htbp]
\centering
\begin{tabular}{cccccc}
\toprule
Label 1 & Label 2 & $\hat{y}_2 - \hat{y}_1$ & Lower bound & Upper bound & p-value \\
\midrule
+ & - & 0.0647 & -0.4346 & 0.5640 & 0.9501 \\
+ & = & -0.1918 & -0.6919 & 0.3084 & 0.6395 \\
- & = & -0.2565 & -0.7557 & 0.2428 & 0.4491 \\
\bottomrule
\end{tabular}
\caption{Pairwise Tukey test for Tone question}
\label{tab:overall_tukey_tone}
\end{table}

\begin{table}[htbp]
\centering
\begin{tabular}{cccccc}
\toprule
Label 1 & Label 2 & $\hat{y}_2 - \hat{y}_1$ & Lower bound & Upper bound & p-value \\
\midrule
+ & - & 0.3283 & -0.1849 & 0.8415 & 0.2898 \\
+ & = & 0.1027 & -0.4113 & 0.6168 & 0.8854 \\
- & = & -0.2256 & -0.7388 & 0.2876 & 0.5560 \\
\bottomrule
\end{tabular}
\caption{Pairwise Tukey test for Clarity question}
\label{tab:overall_tukey_clarity}
\end{table}

\begin{table}[htbp]
\centering
\begin{tabular}{cccccc}
\toprule
Label 1 & Label 2 & $\hat{y}_2 - \hat{y}_1$ & Lower bound & Upper bound & p-value \\
\midrule
+ & - & 0.2934 & -0.2186 & 0.8054 & 0.3696 \\
+ & = & 0.0479 & -0.4649 & 0.5608 & 0.9737 \\
- & = & -0.2455 & -0.7574 & 0.2665 & 0.4976 \\
\bottomrule
\end{tabular}
\caption{Pairwise Tukey test for Intent question}
\label{tab:overall_tukey_intent}
\end{table}

\textbf{Per-topic Evaluation:} Just as with the broad cross-topic evaluation, the performed Levene's tests (\Cref{app:per_topic_plots}) show that observed results display homoscedasticity. The one potential exception to this is for the results of the Intent question for Entertainment topic messages, where $p=0.050$. However, In this case the range of response values was very small, stretching having lower bounds of only 6 or 7 depending on the label, causing relatively small differences in response (when considering the range of possible values) to represent large variations. Even so, across all questions, we fail to find statistically significant evidence that the provided label (or lack thereof) induced changes in the perceived clarity, tone, and ability to convey intent of the analyzed messages on a per-topic basis. The numeric results of the performed Tukey's tests are shown in \Cref{app:per_topic_results}. Similarly to our overall evaluation results, we gain see that, despite failing to find significant evidence that different labels alter message perceptions with respect to message topic, we do observe that the mean perceived tone, clarity, and ability to convey intent are all nearly always lower for messages labeled as having AI assistance (\texttt{+}) compared to messages labeled as \textit{not} having AI assistance (\texttt{-}). The sole exceptions to this are tone for Gossip messages, ability to convey intent for Informational messages, and for every feature of the Recommendation messages. Thus an effect still may be present, despite our failure to find significant to validate it. This again suggests that a larger sample size and/or better granularity in participant quantifications of tone, clarity, and ability to convey intent may draw out this effect. However, the presence of counterfactual instances provides similar evidence that such observed behaviors may just as easily be anomalous.

%% file: discussion.tex
\section{Discussion}

Across all performed analysis, we failed to find statistically significant evidence that the displayed label induced any change in how message recipients perceived the tone, clarity, or ability to convey intent of the messages, independent of message topic. Interestingly, this proves counter to current intuitions and previous literature. Specifically, our analysis fails to align broadly with the observations of \cite{hohensteinArtificialIntelligenceCommunication2023,liuWillAIConsole2022,jakeschAIMediatedCommunicationHow2019}, which all found that users generally perceived content more negatively when told it had been AI-assisted/generated. This suggests that, contrary to prior expectations, the belief that a text message was created with or without AI assistance does not impact how that message is perceived by the recipient with respect to tone, clarity, and intent. To further justify our observations, we point to the conclusions of \citet{jakeschAIMediatedCommunicationHow2019}, which found that the ubiquitous usage of AI induced similar trustworthiness to ubiquitous non-usage. With this in mind, a potential explanation for the observed effect may be that the usage of AI chatbots and LLMs (such as ChatGPT) may have become so ubiquitous within the last year~\cite{shoufanExploringStudentsPerceptionsChatGPT} as to induce the effect observed by \cite{jakeschAIMediatedCommunicationHow2019} more broadly into general life.

However, it is important to note that these results may also be impacted by a few specific features of our study. Firstly, because the vast majority of participants were university students in STEM fields, it is possible that a sampling bias is present that is favorable towards AI presence compared to the broader population. This is because generative AI and LLMs are highly used and positively regarded by university students~\cite{shoufanExploringStudentsPerceptionsChatGPT}. If a larger population were sampled, particularly including (non-STEM) educators and lay people, different outcomes may be observed. That said, if the stated sampling bias is present, such behaviors may in the future extend to the broader population as understanding and usage of new AI techniques improve. That is, the observed behavior may be resultant from an ``early adopter'' effect, which will either decay as users become disenchanted with the technology, or become present in the broader population as time goes on and usage increases.

Secondly, the style in which labels were applied and messages were displayed to participants may also have a subtle impact on the results. The design choices made in this study regarding how messages were composed, labeled, and displayed were primarily made to balance realism with efficient survey completion while avoiding inducing artificial biases. In particular, how the labels are phrased (what words are used to describe the AI assistance or lack  thereof) or displayed (e.g., if one of the class's labels were red instead of both being blue) may induce different results. Additionally, in this study, participants were only shown a single conversation-initiating message without any context. In practice, the context in which a message is received, such as prior history with the sender or if the message was in the middle of a conversation, may potentially impact results. 

Lastly, the questions asked to participants are relatively simple and do not encompass all aspects of human conversational perceptions. Our analysis only focused on tone, clarity, and ability to convey intent. However, features like urgency, general interpretation, terseness, emotionality, etc., also ought to be analyzed. Further, within the features we did analyze, the usage of a simple 0--10 Likert scale may be unable to encode more nuanced differences in perceptions due to the provided labels. As an alternative design, different observations may be observed by directly comparing messages with each on the basis of the analyzed characteristics, rather than analyzed individually (e.g., ``between message A and message B, which is clearer?'' where the two messages may have the same or different labels, and then comparing the average rankings).

Overall, our results provide notable evidence to suggest that the development of LLM-based text-composition assistance for AI-MC can greatly improve text communication without harming message perception. The previously-identified benefits of AI-MC, LLM-based chatbots, and other AI-based assistance tools towards communication and other tasks greatly justify the expansion and improvement of AI-assistance in text communications~\cite{hohensteinArtificialIntelligenceCommunication2023,valenciaLessTypeBetter2023,fuTextSelfUsers2023}.

%% file: conclusion.tex
\section{Conclusion}

In this study, we evaluated whether the labeled presence of AI assistance in text communication induced differences in how message recipients perceived messages. In light of recent rapid developments in generative AI, we aim to guide future developments in AI-MC and AI-assisted text composition with respect to potential unintended counter-effects of AI usage. Through our empirical analysis, we find that the belief by the recipient that AI was utilized to assist in text message composition does not have a significant impact on the perceived tone, clarity, or ability to convey intent of the original message. This indicates that, at least for the surveyed population (primarily university students in STEM fields), we lack evidence to suggest that labeling the usage of AI would harm perceptions of text messages. This is an exciting finding justifies future usage of AI to aid in communication tasks which many people may otherwise find difficult or stressful.

These findings provide foundation for future developments in AI-assisted text composition and AI-MC. To expand practically on this study, future work should consider the implementation of an on-the-fly AI text composition assistant and evaluate how text composers and recipients perceive its usage. In particular, such AI assistance may greatly improve accuracy, speed, and confidence of message composition. Additionally, future research may aim to study the long-term impact of the usage of AI-assisted text composition on human emotional factors, such as sociability, anxiety, stress, satisfaction, and mood. Alternatively, future works may desire to replicate this study using a larger and more varied sample of the population to build a broader understanding of the effect, if any, that believed AI usage has on human message perception.

%% file: appendix.tex
\section{Per-Topic Test Plots}
\label{app:per_topic_plots}

\begin{figure}[htbp]
\centering
\begin{subfigure}{.5\linewidth}
    \centering
    \includegraphics[width=\linewidth]{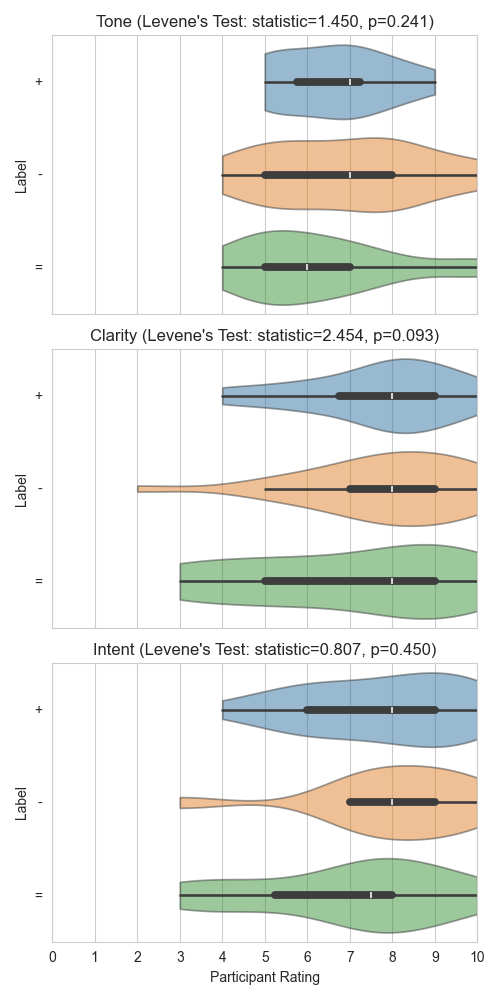}
    \caption{Violin plots with Levene's test metrics}
\end{subfigure}%
\begin{subfigure}{.5\linewidth}
    \centering
    \includegraphics[width=\linewidth]{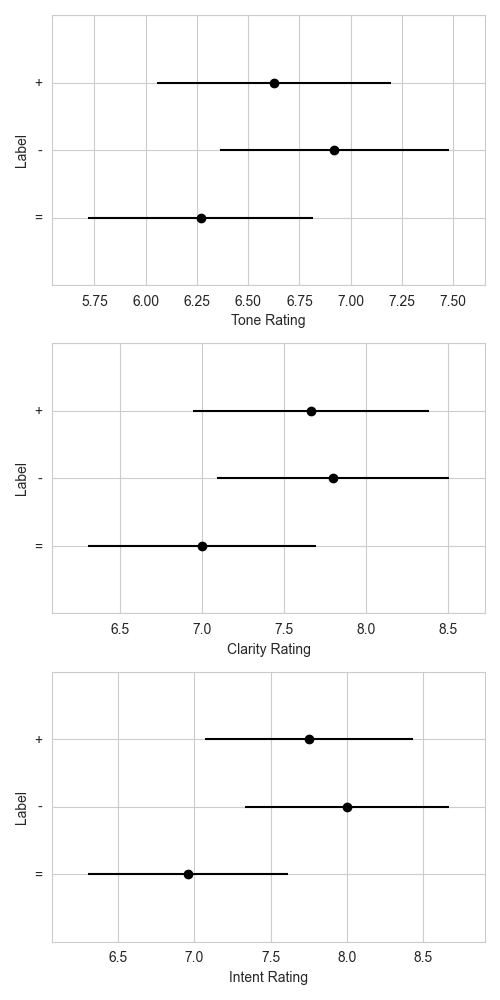}
    \caption{Interval plots for Tukey's test}
\end{subfigure}
\caption{Plots for performed statistical tests on Advice messages}
\end{figure}

\begin{figure}[htbp]
\centering
\begin{subfigure}{.5\linewidth}
    \centering
    \includegraphics[width=\linewidth]{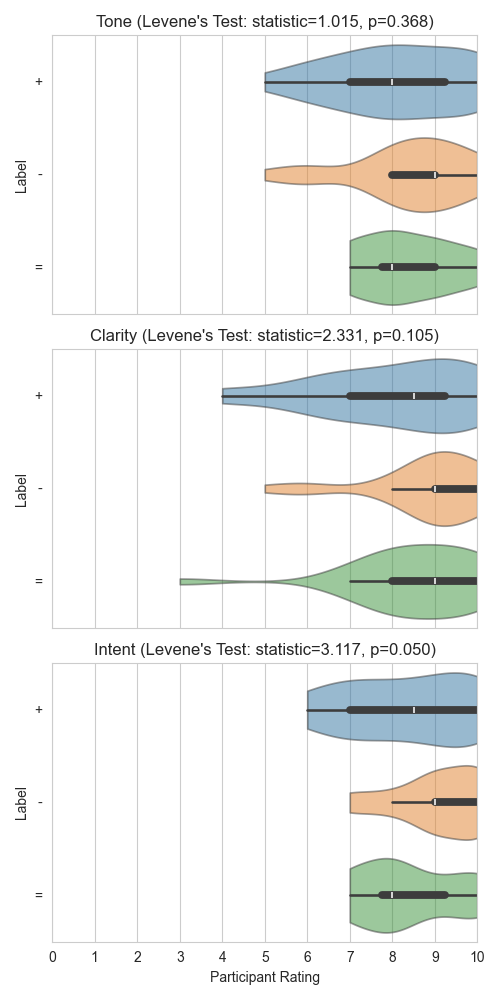}
    \caption{Violin plots with Levene's test metrics}
\end{subfigure}%
\begin{subfigure}{.5\linewidth}
    \centering
    \includegraphics[width=\linewidth]{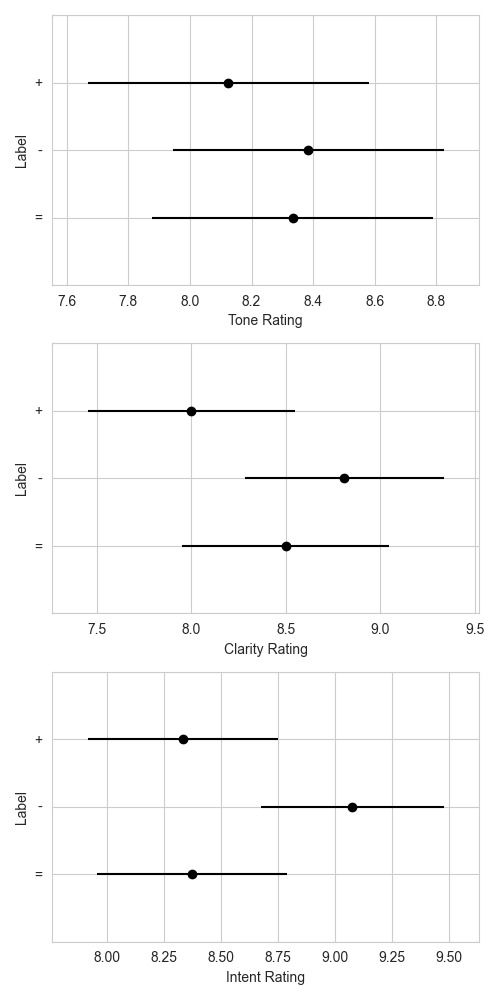}
    \caption{Interval plots for Tukey's test}
\end{subfigure}
\caption{Plots for performed statistical tests on Entertainment messages}
\end{figure}

\begin{figure}[htbp]
\centering
\begin{subfigure}{.5\linewidth}
    \centering
    \includegraphics[width=\linewidth]{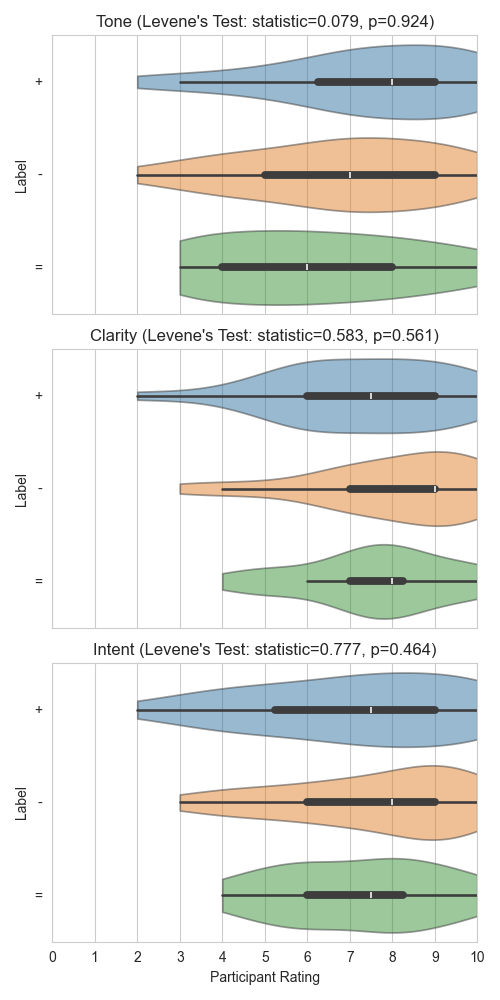}
    \caption{Violin plots with Levene's test metrics}
\end{subfigure}%
\begin{subfigure}{.5\linewidth}
    \centering
    \includegraphics[width=\linewidth]{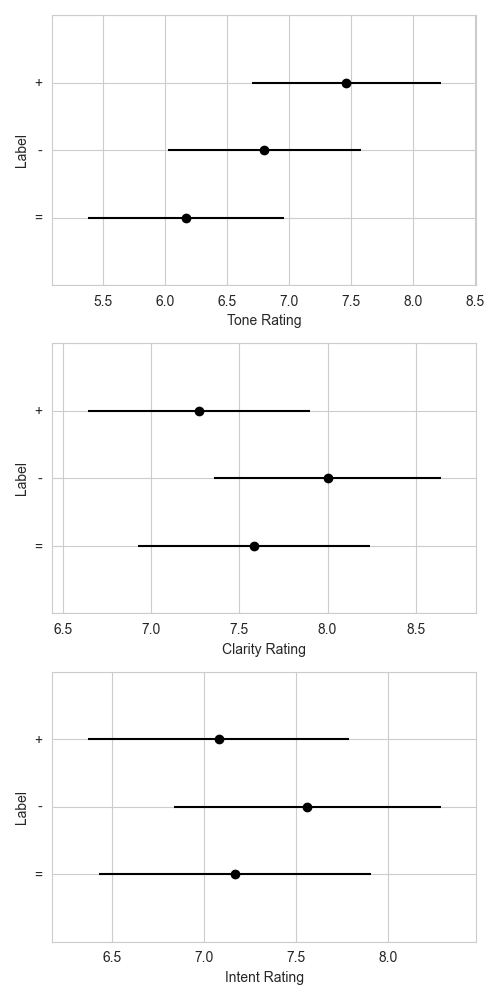}
    \caption{Interval plots for Tukey's test}
\end{subfigure}
\caption{Plots for performed statistical tests on Gossip messages}
\end{figure}

\begin{figure}[htbp]
\centering
\begin{subfigure}{.5\linewidth}
    \centering
    \includegraphics[width=\linewidth]{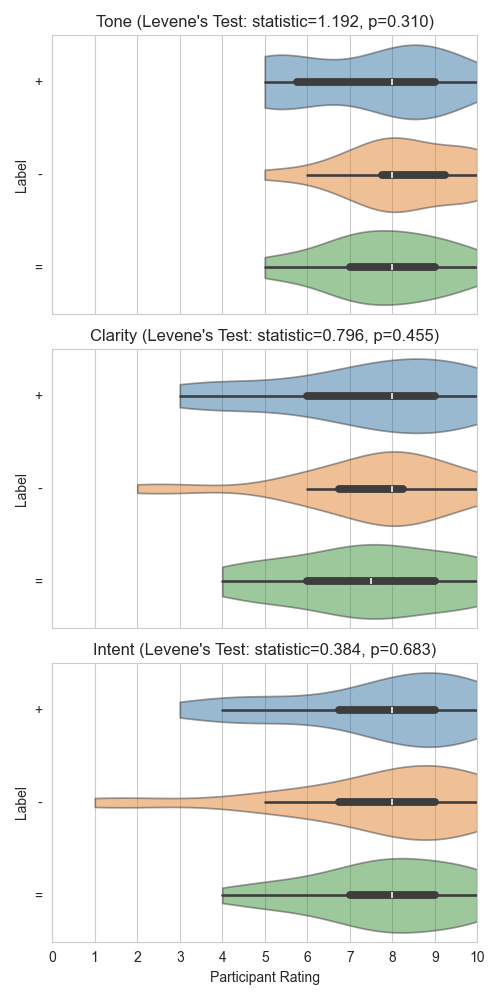}
    \caption{Violin plots with Levene's test metrics}
\end{subfigure}%
\begin{subfigure}{.5\linewidth}
    \centering
    \includegraphics[width=\linewidth]{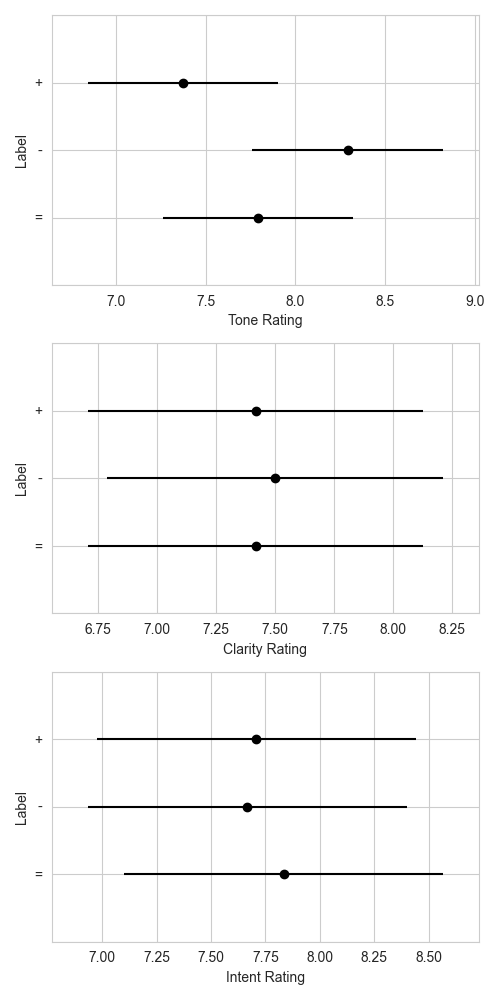}
    \caption{Interval plots for Tukey's test}
\end{subfigure}
\caption{Plots for performed statistical tests on Informational messages}
\end{figure}

\begin{figure}[htbp]
\centering
\begin{subfigure}{.5\linewidth}
    \centering
    \includegraphics[width=\linewidth]{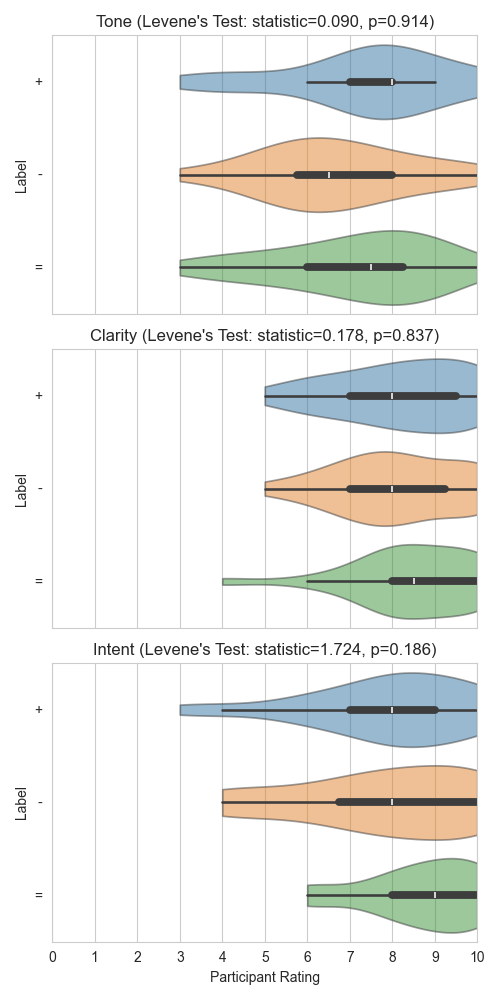}
    \caption{Violin plots with Levene's test metrics}
\end{subfigure}%
\begin{subfigure}{.5\linewidth}
    \centering
    \includegraphics[width=\linewidth]{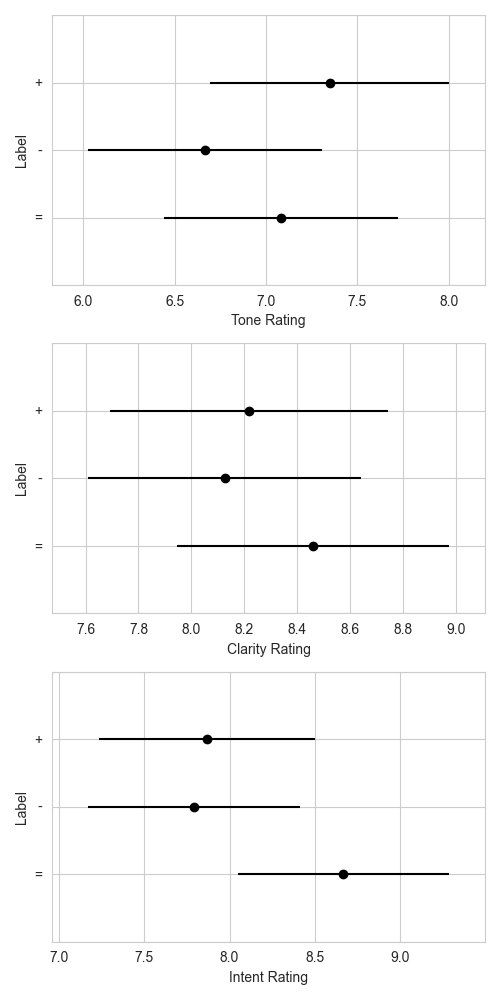}
    \caption{Interval plots for Tukey's test}
\end{subfigure}
\caption{Plots for performed statistical tests on Recommendation messages}
\end{figure}

\begin{figure}[htbp]
\centering
\begin{subfigure}{.5\linewidth}
    \centering
    \includegraphics[width=\linewidth]{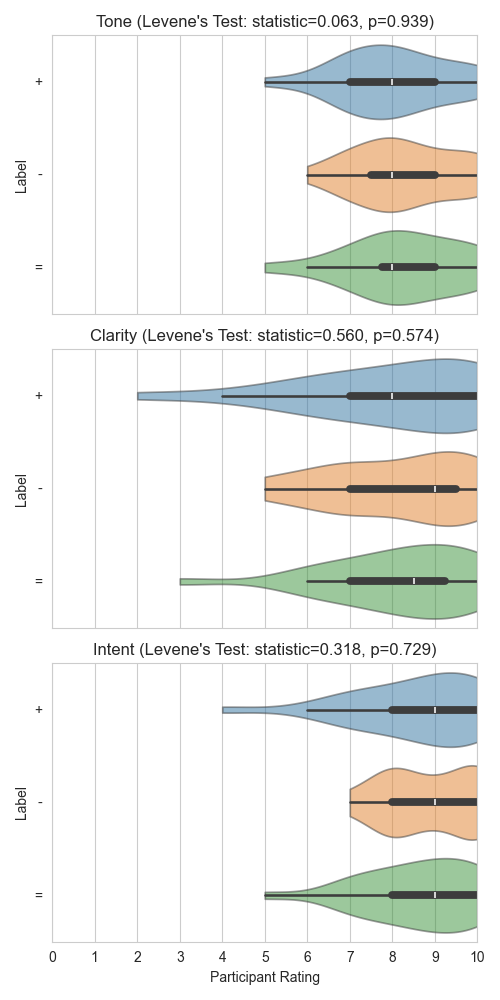}
    \caption{Violin plots with Levene's test metrics}
\end{subfigure}%
\begin{subfigure}{.5\linewidth}
    \centering
    \includegraphics[width=\linewidth]{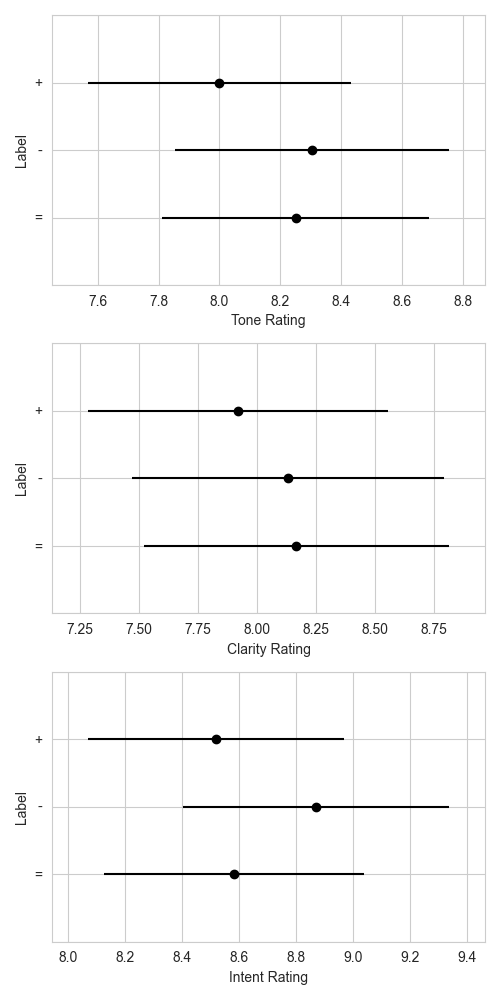}
    \caption{Interval plots for Tukey's test}
\end{subfigure}
\caption{Plots for performed statistical tests on Social messages}
\end{figure}

\clearpage

\section{Per-Topic Tukey Test Results}
\label{app:per_topic_results}

\begin{table}[htbp]
\centering
\caption{Pairwise Tukey test for Tone question on Advice messages}
\begin{tabular}{cccccc}
\toprule
Label 1 & Label 2 & $\hat{y}_2 - \hat{y}_1$ & Lower bound & Upper bound & p-value \\
\midrule
+ & - & 0.2950 & -0.8350 & 1.4250 & 0.8071 \\
+ & = & -0.3558 & -1.4751 & 0.7636 & 0.7282 \\
- & = & -0.6508 & -1.7584 & 0.4569 & 0.3432 \\
\bottomrule
\end{tabular}
\end{table}

\begin{table}[htbp]
\centering
\caption{Pairwise Tukey test for Clarity question on Advice messages}
\begin{tabular}{cccccc}
\toprule
Label 1 & Label 2 & $\hat{y}_2 - \hat{y}_1$ & Lower bound & Upper bound & p-value \\
\midrule
+ & - & 0.1333 & -1.2933 & 1.5600 & 0.9728 \\
+ & = & -0.6667 & -2.0798 & 0.7465 & 0.4995 \\
- & = & -0.8000 & -2.1984 & 0.5984 & 0.3624 \\
\bottomrule
\end{tabular}
\end{table}

\begin{table}[htbp]
\centering
\caption{Pairwise Tukey test for Intent question on Advice messages}
\begin{tabular}{cccccc}
\toprule
Label 1 & Label 2 & $\hat{y}_2 - \hat{y}_1$ & Lower bound & Upper bound & p-value \\
\midrule
+ & - & 0.2500 & -1.0974 & 1.5974 & 0.8972 \\
+ & = & -0.7885 & -2.1231 & 0.5461 & 0.3392 \\
- & = & -1.0385 & -2.3591 & 0.2822 & 0.1513 \\
\bottomrule
\end{tabular}
\end{table}

\begin{table}[htbp]
\centering
\caption{Pairwise Tukey test for Tone question on Entertainment messages}
\begin{tabular}{cccccc}
\toprule
Label 1 & Label 2 & $\hat{y}_2 - \hat{y}_1$ & Lower bound & Upper bound & p-value \\
\midrule
+ & - & 0.2596 & -0.6363 & 1.1555 & 0.7679 \\
+ & = & 0.2083 & -0.7053 & 1.1220 & 0.8489 \\
- & = & -0.0513 & -0.9472 & 0.8446 & 0.9897 \\
\bottomrule
\end{tabular}
\end{table}

\begin{table}[htbp]
\centering
\caption{Pairwise Tukey test for Clarity question on Entertainment messages}
\begin{tabular}{cccccc}
\toprule
Label 1 & Label 2 & $\hat{y}_2 - \hat{y}_1$ & Lower bound & Upper bound & p-value \\
\midrule
+ & - & 0.8077 & -0.2653 & 1.8807 & 0.1763 \\
+ & = & 0.5000 & -0.5943 & 1.5943 & 0.5210 \\
- & = & -0.3077 & -1.3807 & 0.7653 & 0.7722 \\
\bottomrule
\end{tabular}
\end{table}

\begin{table}[htbp]
\centering
\caption{Pairwise Tukey test for Intent question on Entertainment messages}
\begin{tabular}{cccccc}
\toprule
Label 1 & Label 2 & $\hat{y}_2 - \hat{y}_1$ & Lower bound & Upper bound & p-value \\
\midrule
+ & - & 0.7436 & -0.0718 & 1.5590 & 0.0811 \\
+ & = & 0.0417 & -0.7899 & 0.8732 & 0.9921 \\
- & = & -0.7019 & -1.5173 & 0.1135 & 0.1055 \\
\bottomrule
\end{tabular}
\end{table}

\begin{table}[htbp]
\centering
\caption{Pairwise Tukey test for Tone question on Gossip messages}
\begin{tabular}{cccccc}
\toprule
Label 1 & Label 2 & $\hat{y}_2 - \hat{y}_1$ & Lower bound & Upper bound & p-value \\
\midrule
+ & - & -0.6615 & -2.1971 & 0.8740 & 0.5599 \\
+ & = & -1.2949 & -2.8466 & 0.2569 & 0.1203 \\
- & = & -0.6333 & -2.1999 & 0.9332 & 0.5997 \\
\bottomrule
\end{tabular}
\end{table}

\begin{table}[htbp]
\centering
\caption{Pairwise Tukey test for Clarity question on Gossip messages}
\begin{tabular}{cccccc}
\toprule
Label 1 & Label 2 & $\hat{y}_2 - \hat{y}_1$ & Lower bound & Upper bound & p-value \\
\midrule
+ & - & 0.7308 & -0.5405 & 2.0020 & 0.3590 \\
+ & = & 0.3141 & -0.9706 & 1.5988 & 0.8285 \\
- & = & -0.4167 & -1.7136 & 0.8803 & 0.7232 \\
\bottomrule
\end{tabular}
\end{table}

\begin{table}[htbp]
\centering
\caption{Pairwise Tukey test for Intent question on Gossip messages}
\begin{tabular}{cccccc}
\toprule
Label 1 & Label 2 & $\hat{y}_2 - \hat{y}_1$ & Lower bound & Upper bound & p-value \\
\midrule
+ & - & 0.4831 & -0.9549 & 1.9211 & 0.7018 \\
+ & = & 0.0897 & -1.3635 & 1.5430 & 0.9880 \\
- & = & -0.3933 & -1.8604 & 1.0738 & 0.7977 \\
\bottomrule
\end{tabular}
\end{table}

\begin{table}[htbp]
\centering
\caption{Pairwise Tukey test for Tone question on Informational messages}
\begin{tabular}{cccccc}
\toprule
Label 1 & Label 2 & $\hat{y}_2 - \hat{y}_1$ & Lower bound & Upper bound & p-value \\
\midrule
+ & - & 0.9167 & -0.1446 & 1.9779 & 0.1039 \\
+ & = & 0.4167 & -0.6446 & 1.4779 & 0.6168 \\
- & = & -0.5000 & -1.5613 & 0.5613 & 0.5000 \\
\bottomrule
\end{tabular}
\end{table}

\begin{table}[htbp]
\centering
\caption{Pairwise Tukey test for Clarity question on Informational messages}
\begin{tabular}{cccccc}
\toprule
Label 1 & Label 2 & $\hat{y}_2 - \hat{y}_1$ & Lower bound & Upper bound & p-value \\
\midrule
+ & - & 0.0833 & -1.3383 & 1.5050 & 0.9892 \\
+ & = & 0.0000 & -1.4216 & 1.4216 & 1.0000 \\
- & = & -0.0833 & -1.5050 & 1.3383 & 0.9892 \\
\bottomrule
\end{tabular}
\end{table}

\begin{table}[htbp]
\centering
\caption{Pairwise Tukey test for Intent question on Informational messages}
\begin{tabular}{cccccc}
\toprule
Label 1 & Label 2 & $\hat{y}_2 - \hat{y}_1$ & Lower bound & Upper bound & p-value \\
\midrule
+ & - & -0.0417 & -1.5064 & 1.4231 & 0.9974 \\
+ & = & 0.1250 & -1.3398 & 1.5898 & 0.9772 \\
- & = & 0.1667 & -1.2981 & 1.6314 & 0.9599 \\
\bottomrule
\end{tabular}
\end{table}

\begin{table}[htbp]
\centering
\caption{Pairwise Tukey test for Tone question on Recommendation messages}
\begin{tabular}{cccccc}
\toprule
Label 1 & Label 2 & $\hat{y}_2 - \hat{y}_1$ & Lower bound & Upper bound & p-value \\
\midrule
+ & - & -0.6812 & -1.9737 & 0.6113 & 0.4210 \\
+ & = & -0.2645 & -1.5570 & 1.0280 & 0.8761 \\
- & = & 0.4167 & -0.8620 & 1.6953 & 0.7160 \\
\bottomrule
\end{tabular}
\end{table}

\begin{table}[htbp]
\centering
\caption{Pairwise Tukey test for Clarity question on Recommendation messages}
\begin{tabular}{cccccc}
\toprule
Label 1 & Label 2 & $\hat{y}_2 - \hat{y}_1$ & Lower bound & Upper bound & p-value \\
\midrule
+ & - & -0.0924 & -1.1325 & 0.9477 & 0.9753 \\
+ & = & 0.2409 & -0.7992 & 1.2811 & 0.8442 \\
- & = & 0.3333 & -0.6957 & 1.3623 & 0.7188 \\
\bottomrule
\end{tabular}
\end{table}

\begin{table}[htbp]
\centering
\caption{Pairwise Tukey test for Intent question on Recommendation messages}
\begin{tabular}{cccccc}
\toprule
Label 1 & Label 2 & $\hat{y}_2 - \hat{y}_1$ & Lower bound & Upper bound & p-value \\
\midrule
+ & - & -0.0779 & -1.3295 & 1.1737 & 0.9878 \\
+ & = & 0.7971 & -0.4545 & 2.0487 & 0.2853 \\
- & = & 0.8750 & -0.3632 & 2.1132 & 0.2152 \\
\bottomrule
\end{tabular}
\end{table}

\begin{table}[htbp]
\centering
\caption{Pairwise Tukey test for Tone question on Social messages}
\begin{tabular}{cccccc}
\toprule
Label 1 & Label 2 & $\hat{y}_2 - \hat{y}_1$ & Lower bound & Upper bound & p-value \\
\midrule
+ & - & 0.3043 & -0.5749 & 1.1836 & 0.6863 \\
+ & = & 0.2500 & -0.6196 & 1.1196 & 0.7709 \\
- & = & -0.0543 & -0.9423 & 0.8336 & 0.9882 \\
\bottomrule
\end{tabular}
\end{table}

\begin{table}[htbp]
\centering
\caption{Pairwise Tukey test for Clarity question on Social messages}
\begin{tabular}{cccccc}
\toprule
Label 1 & Label 2 & $\hat{y}_2 - \hat{y}_1$ & Lower bound & Upper bound & p-value \\
\midrule
+ & - & 0.2104 & -1.0851 & 1.5059 & 0.9200 \\
+ & = & 0.2467 & -1.0347 & 1.5280 & 0.8896 \\
- & = & 0.0362 & -1.2721 & 1.3446 & 0.9976 \\
\bottomrule
\end{tabular}
\end{table}

\begin{table}[htbp]
\centering
\caption{Pairwise Tukey test for Intent question on Social messages}
\begin{tabular}{cccccc}
\toprule
Label 1 & Label 2 & $\hat{y}_2 - \hat{y}_1$ & Lower bound & Upper bound & p-value \\
\midrule
+ & - & 0.3496 & -0.5657 & 1.2648 & 0.6328 \\
+ & = & 0.0633 & -0.8419 & 0.9686 & 0.9846 \\
- & = & -0.2862 & -1.2106 & 0.6381 & 0.7396 \\
\bottomrule
\end{tabular}
\end{table}